\documentclass[conf]{new-aiaa}

\usepackage[utf8]{inputenc}

\usepackage{graphicx}
\usepackage{wrapfig}
\usepackage{amsmath}
\usepackage{comment}
\usepackage[version=4]{mhchem}
\usepackage{siunitx}
\usepackage{longtable,tabularx}
\usepackage{fullpage}
\usepackage{url}
\usepackage{todonotes}
\usepackage{multirow}
\usepackage{makecell}
\usepackage{hyperref}
\usepackage{bm}
\usepackage{algorithm2e}
\usepackage{indentfirst}

\usepackage{caption}
\usepackage{subcaption}
\usepackage{tikz}
\usetikzlibrary{arrows,automata,%
                arrows.meta,%
                petri,%
                topaths}
                
\newcolumntype{L}{>{\centering\arraybackslash}m{4cm}}
\newcolumntype{M}{>{\centering\arraybackslash}m{1.5cm}}
\usepackage{xcolor}
\newif\ifdraft
\draftfalse

\newcommand{\jiahong}[1]{\ifdraft{\color{magenta}[{\bf Jiahong}: #1]}\fi}

\newcommand{\jb}[1]{\ifdraft{\color{red}[{\bf JB}: #1]}\fi}

\newcommand\ft{{\ensuremath\,\rm ft}}

\newcommand\fpm{{\ensuremath\,\rm fpm}}
\newcommand\kt{{\ensuremath\,\rm kt}}
\newcommand\kts{{\ensuremath\,\rm kts}}
\newcommand\m{{\ensuremath\,\rm m}}
\newcommand\ms{{\ensuremath\,\rm m.s^{-1}}}

\title{Falsification of a Vision-based Automatic Landing System}
\author{Sara Shoouri\footnote{Master's Student, Department of  Electrical Engineering and Computer Science, University of Michigan, Ann Arbor, MI.}}
\author{Shayan  Jalili\footnote{PhD Student, Department of Aerospace Engineering, University of Michigan, Ann Arbor, MI.}}
\author{Jiahong Xu\footnote{Master's Student, Department of  Electrical Engineering and Computer Science, University of Michigan, Ann Arbor, MI.}}
\author{Isabelle  Gallagher\footnote{Undergraduate Student, Department of  Electrical Engineering and Computer Science, University of Michigan, Ann Arbor, MI.}}
\author{Yuhao Zhang\footnote{Master's Student, Department of  Mechanical Engineering, University of Michigan, Ann Arbor, MI.}}
\author{Joshua Wilhelm\footnote{Master's Student, Department of Aerospace Engineering, University of Michigan, Ann Arbor, MI.}}
\author{Necmiye Ozay\footnote{Associate Professor, Department of  Electrical Engineering and Computer Science, University of Michigan, Ann Arbor, MI.}}
\author{Jean-Baptiste~Jeannin\footnote{Assistant Professor, Department of Aerospace Engineering, University of Michigan, Ann Arbor, MI.}}
\affil{University of Michigan, Ann Arbor, MI 48109, U.S.A.}

\begin{document}

\maketitle

\begin{abstract}

At smaller airports without an instrument approach or advanced equipment, automatic landing of aircraft is a safety-critical task that requires the use of sensors present on the aircraft.
In this paper, we study falsification of an automatic landing system for fixed-wing aircraft using a camera as its main sensor.
We first present an architecture for vision-based automatic landing, including a vision-based runway distance and orientation estimator and an associated PID controller.
We then outline landing specifications that we validate with actual flight data.
Using these specifications, we propose the use of the falsification tool Breach to find counterexamples to the specifications in the automatic landing system.
Our experiments are implemented using a Beechcraft Baron 58 in the X-Plane flight simulator communicating with MATLAB Simulink.
\end{abstract}


\section{Introduction}


Single-pilot operations and fully automatic flight for large commercial aircraft promise cheaper air transportation for both passengers and cargo. 
One remaining technical challenge is to enable automatic landing at all airports.
%
Automatic, pilot-supervised landings for large aircraft such as airliners have been available since the late 1960s with the invention of the Autoland system~\cite{charnley1959blind}.
However, these systems require not only a qualified aircraft and crew, but also very expensive equipment on the ground in the form of an Instrument Landing System (ILS), often Category II or Category III.
As a result, Autoland is only available at the biggest airports in the world, or at airports which most often have bad weather. 
With the recent interest in single-pilot and unmanned operations for commercial aircraft, there is an increased interest in expanding automatic landing operations to smaller airports without special equipments and often no Instrument Flight Rules (IFR) approach.

In Visual Meteorological Conditions (VMC), pilots operating aircraft mainly rely on their vision to estimate the position and safely land on the runway.
This paper explores how to exploit this intuition to create an automatic landing system that uses a camera as a sensor, and vision algorithms to identify the runway on the images, estimate the position of the aircraft with respect to the runway, and safely control the aircraft to a landing.
In addition to the camera, the algorithm can also use instrument data from the airplane such as Global Positioning System (GPS), gyroscopes, and velocity measurements.
We use vision as a primary sensor because the GPS signal is typically not precise enough (especially in terms of the altitude of the aircraft) to enable a fully automatic landing on its own, and it would also require expensive calibration at the airport.
A prototype of the system was implemented using MATLAB Simulink connected to the X-Plane flight simulator via the X-Plane Connect Toolbox \cite{xplaneconnect}. Results are demonstrated on a Baron 58 at Ann Arbor Airport.

\begin{figure}[h]
\centering
\includegraphics[width=1\textwidth]{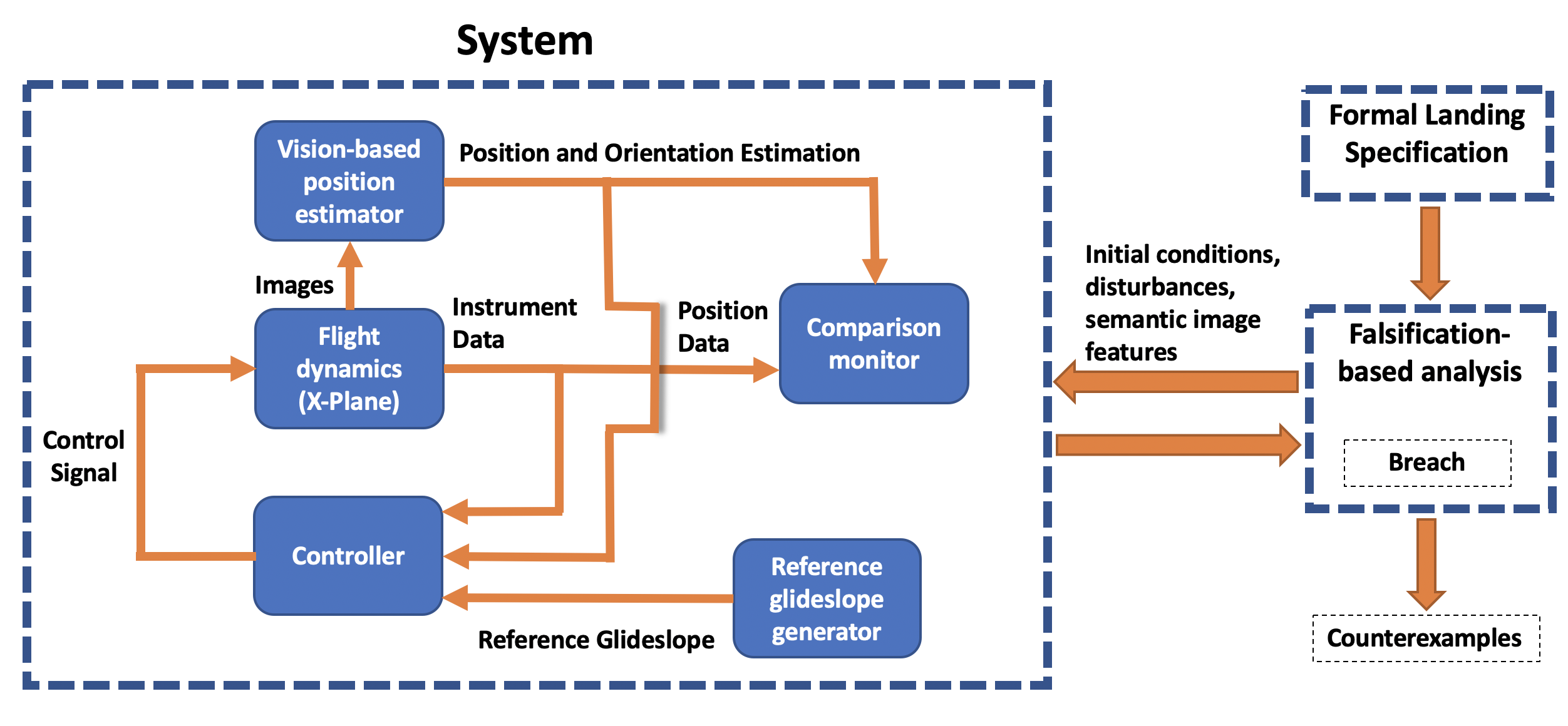}
\caption{The proposed vision-based automatic landing system is shown in the system box. Its interface with the falsification tool Breach for analysis purposes is also depicted. 
}
\label{fig:framework}
\end{figure}

Automatic landing systems are safety-critical for the operation of the controlled aircraft, and the algorithms they use must be carefully verified. Our primary interest in designing this automatic landing system is to run falsification algorithms to find potential counterexamples in the design or implementation.
As a first step, we precisely specify the requirements of an automatic landing system using linear temporal logic~\cite{pnueli1977temporal},
and we validate those specifications using two different flights, one airline flight and one private flight.
Once the requirements are specified, the falsification tool Breach~\cite{donze2010breach} is run to find counter-examples to either the image processing algorithms or the tracking controllers.
The proposed vision-based automatic landing system and its interface with the falsification tool is shown in Figure \ref{fig:framework}.

Instrument-based automatic landing systems have been available since the 1960s~\cite{charnley1959blind}, but  vision-based methods for automatic landing of fixed-wing UAVs have only been proposed recently~\cite{huh2010vision,laiacker2013vision,min2007guidance,thurrowgood2014biologically,le2009nonlinear,
burlion2014backstepping,coutard2011automatic,bourquardez2007visual,victor2013landing,rives2002visual}.
To the best of our knowledge, falsification or verification techniques have not been previously employed on any of these systems, and verification work has focused on landing protocols rather than full systems~\cite{oishi2002hybrid,umeno2007safety,johnson2012parametrized}.
On the industrial side, Garmin is currently developing an emergency autoland feature for small general aviation aircraft~\cite{george2019flying}.
Additionally, a similar architecture to the one presented in this paper targets automatic taxiing~\cite{zhang2020software}, where a vision algorithm was originally part of the proposed architecture, but was not implemented or analyzed.




\section{Automatic Landing Architecture}
A prototype vision-based automatic landing architecture is developed as shown in Figure~\ref{fig:framework}  inside the System block. We use X-Plane as the aircraft simulator, which can provide both data from flight instruments and real-time video feed from cameras located at desired positions on the aircraft. The architecture consists of a perception module for processing the camera data, a reference glideslope generator which creates waypoints along the approach path, and a feedback controller that aims to generate control inputs in order to track the waypoints.  
Specific architecture components for automatic landing and the results after implementation are discussed in the following sections.



\subsection{Vision-based Runway Distance And Orientation Estimation} \label{subsec:vision}

In this section, we describe our vision-based position and orientation estimation algorithms. Camera pose and aircraft pose are used interchangeably as we assume a single camera attached to the tail of the aircraft. As a result, there is a known constant transformation between the two poses. Our proposed vision pipeline consists of three main modules: (i) rough localization and masking of the runway in the image; (ii) precise detection of the pixel coordinates of the runway landmark points on the image; and (iii) estimation of the camera pose based on the pixel coordinates of the runway landmarks, the camera parameters, and the known 3D world coordinates of the landmarks.

\begin{figure}[!th]
\centering
\includegraphics[width=0.9\textwidth]{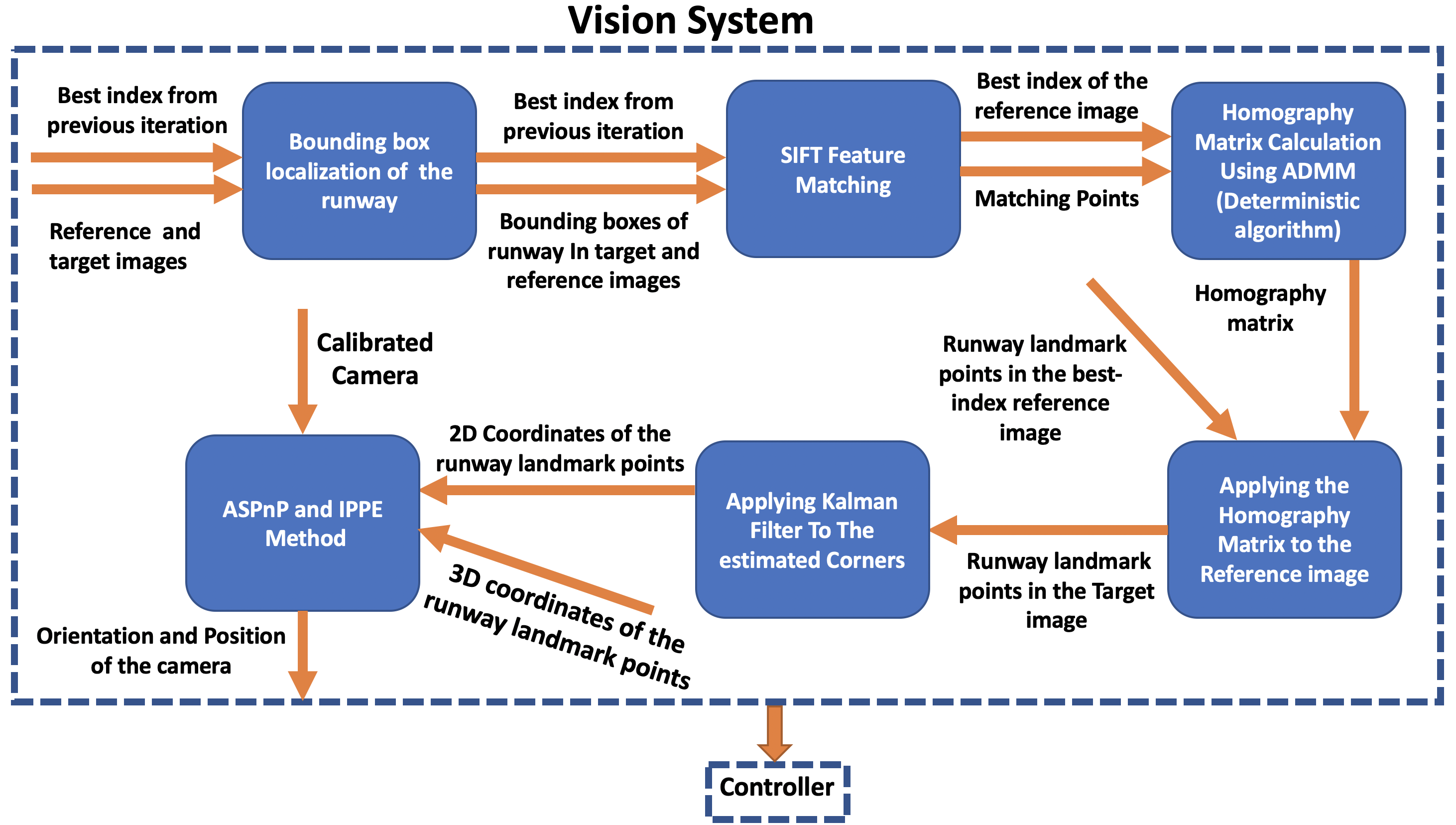}
\caption{The proposed vision pipeline.
}
\label{fig:framework-vision1}
\end{figure}

Our vision-based runway location and orientation estimator is a modular vision pipeline -- as opposed to, e.g., an end-to-end neural network -- and uses several well-known techniques and algorithms:
\begin{itemize}
    \item \emph{YOLO Neural Network:}
    YOLO \cite{Yolo} is a neural network for real-time object detection that determines particular objects' locations in the image. YOLO creates a one-step process, applies the model to images at multiple locations and scales, and then considers the boxes with high scoring as the detected bounding boxes.
    \item \emph{SIFT and Matching points in two images:}
    SIFT \cite{lowe2004distinctive} is a feature detection algorithm to detect local features from the images. A set of keypoints is extracted from the detected SIFT features for two different images.
    The matching feature points are detected based on comparing each feature from the first image with the second image and minimizing the Euclidean distance of the points' feature vectors.
  

    \item \emph{ASPnP and IPPE:}
    The goal of the Perspective-n-Point (PnP) problem is to determine the position and orientation of the camera given its intrinsic matrix and a set of $n$ points with their 2D and 3D coordinates. 
    ASPnP \cite{aspnp} and IPPE \cite{ippe} are two recent algorithms that provide a solution to the PnP problem. The main idea of ASPnP is to estimate the orientation and position parameters by directly minimizing a properly defined algebraic error. The main idea of IPPE is to estimate the parameters using the transform about an infinitesimally small region on the surface (by solving a 1st-order partial differential equation).

\end{itemize}
%
%
%
The overall vision pipeline with the corresponding algorithms are shown in Figure \ref{fig:framework-vision1}, and we explain the details next.

\subsubsection{Rough localization and masking of the runway in the image}
Running SIFT on the entire image is time consuming. 
The goal of this step is to improve the computation time for the SIFT method
by making sure it runs only on a small portion of the image that contains the runway. The bounding box of the runway is localized in each frame using YOLO. To reduce the risk of over-fitting and make the process of training faster and more robust, the fine tuning technique from \cite{finetune} is applied. This approach initializes the
network parameters for the target task from the parameters of the
network pre-trained on another related task. 450 images are gathered during the landing approach from the X-plane simulator and the runway's bounding boxes are localized on each picture manually. After collecting the dataset, it is fed into the pre-trained YOLO network and the model is re-trained based on the collected dataset. 
The trained model is able to find the rough localization of the runway. We apply the trained neural network model to both reference and target frames and localize the location of the runway's bounding boxes. Then, we mask out the bounding boxes from both images and perform the next step on them.

\subsubsection{Precise detection of the pixel coordinates of the runway landmark points on the image}
\par
This step aims to estimate the pixel coordinates of runway landmark points on the images. The vision algorithm switches between two different landmarks: corners of the runway and the runway aiming point markings.
When the airplane distance from the runway is between $800$ (m) and $20$ (m), the vision method uses the runway's corners as the landmark points. However, as the airplane gets closer to the runway, the corners of the runway start vanishing. Therefore, the vision algorithm uses the runway aiming points markings, two rectangular markings consisting of a broad white stripe located on the runway, when the airplane distance from the runway is less than $20$ (m). 
\par
In order to detect the exact pixel coordinates of the runway landmark points, we perform image registration against a stack of reference images generated in X-Plane on the ideal approach path, in which the pixel coordinates of the runway's landmark points are manually annotated. Any two pictures of
the same planar surface in space are related by a $3\times3$ matrix, called the homography matrix, which  we use to compute the
location of the landmark points in the test frames \cite{miller2008landing}. Scale-Invariant Feature Transform (SIFT) is utilized to find the reference images' matching points and their corresponding points in the test images. Subsequently, the Alternating Direction Method of Multipliers (ADMM) and Proximal Block Coordinate Descent (BCD) are applied to compute the homography matrix in an efficient and deterministic way \cite{wen2019efficient}. Additionally, the resulting homography matrix from the previous frame is used to warm-start the optimization in the next step. Since the accuracy of landmark points' localization is crucial for the next step, we use a Kalman filter \cite{kalman} to predict the landmarks' future locations. The Kalman filter addresses two distinct scenarios: $1)$ when the landmark points are detected, and they are close enough to the previous detection, the Kalman filter first predicts the states of the landmark points in the current frame and then uses the newly detected coordinates for the landmark points from the vision module to correct the states; and $2)$ when the detected landmark points are far away from the landmark points in the previous frame, the Kalman filtering relies on its prediction for the states and replaces them as the new landmarks' location.

At runtime, for each frame of the landing approach, we need to find the reference image that best matches the incoming frame. In order to achieve this efficiently without needing to search through all reference images, we define a best-index variable, which is the index of the reference image that is most similar to the test frame. The best-index variable is increased periodically every frame. By knowing the best-index for the previous frame, it is not necessary to match the target frame against all reference frames. We assume that the most similar reference image frame has the highest number of SIFT matching points, which produces the most accurate registration. Each incoming image should be compared with the three reference images corresponding to the previous best-index and its two following indices. The best-index is updated with the index of the reference image that has a higher number of SIFT matching points. 

\par
The output of this step is the pixel coordinates of the landmarks on the current frame.

\subsubsection{Estimation of the camera position and orientation based on the pixel coordinates of the runway landmarks}
 In order to be able to estimate the camera position and orientation, the camera located on the airplane must be calibrated. We assume a simple pinhole camera model and estimated its $3\times 4$ intrinsic camera matrix. We obtained the intrinsic camera matrix by annotating a few points on each image and solving a linear least-squares problem. The annotated points have the known world coordinates and the images were taken at various distances from the camera. Given the pixel coordinates of the landmarks on the current frame, which are computed in step 2, with their 3D coordinates in the world and the camera intrinsic matrix, it is possible to estimate the camera position and orientation via the PnP problem. We have tried several existing algorithms to solve the PnP problem. We have decided to use the
ASPnP algorithm to estimate the position and IPPE to estimate the orientation, since ASPnP can perform better than IPPE in position estimation, but it can have more significant errors on the orientation estimation than IPPE. We have combined these two methods to solve the PnP problem. Altitude estimates from vision were less reliable. To address this issue, we obtained the aircraft's height from the onboard barometer and use that value to enhance the estimation of the orientation, altitude, and lateral deviation.

\begin{figure}[!ht]
  \centering
  \begin{minipage}[b]{0.495\textwidth}
    \includegraphics[width=\textwidth]{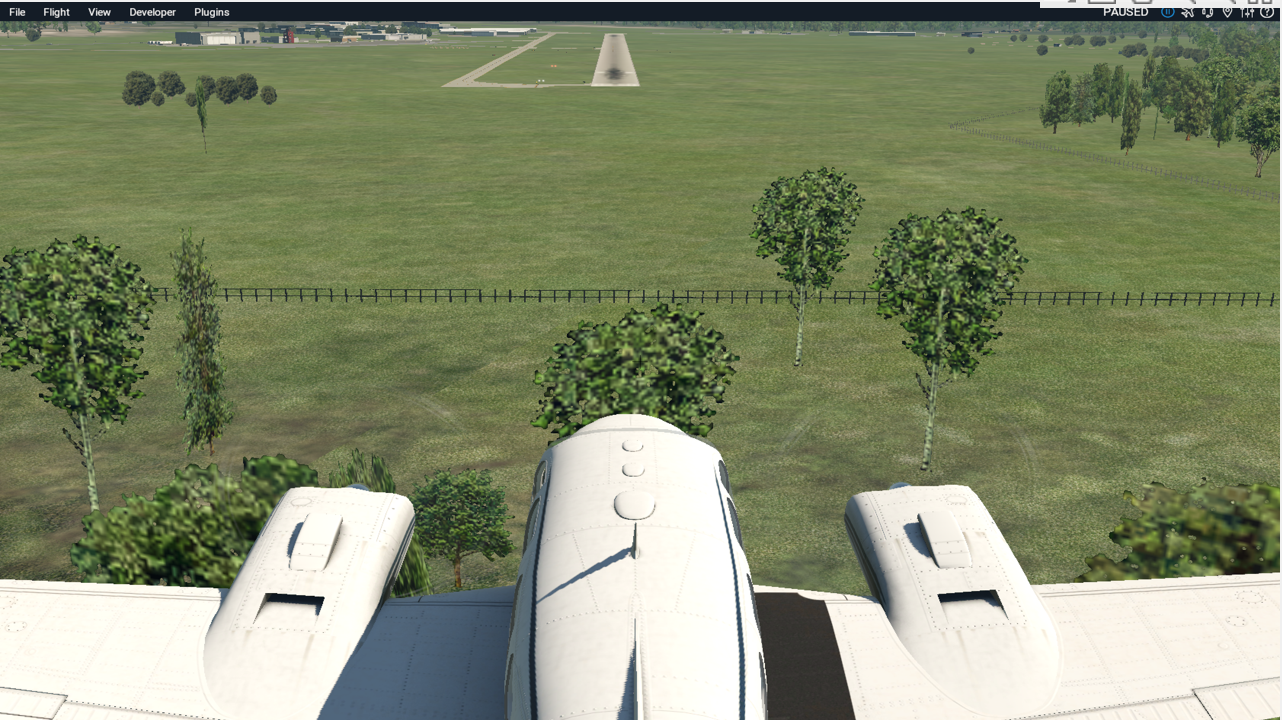}
  \end{minipage}
  \hfill
  \begin{minipage}[b]{0.495\textwidth}
\includegraphics[width=\textwidth]{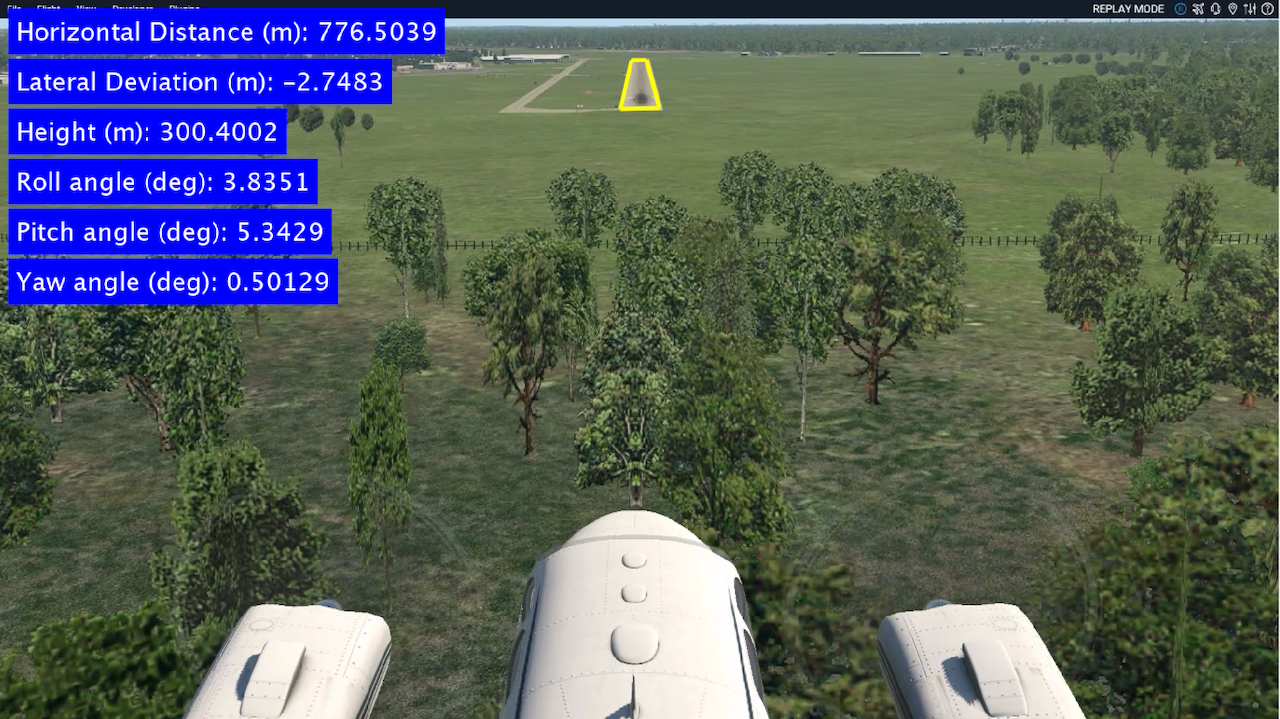}
  \end{minipage}
  \caption{The result of implementing the registration between a reference frame (left) and a test frame (right), and the airplane's position and orientation estimations. The reference image is taken on the glide path. The corners of the runways in the test frame are computed and shown in the picture. The calculated distance of the camera with respect to the runway is shown in the blue box on the top of the test frame.}
  \label{res-vision}
\end{figure}

\vspace{0.5cm}
Figure~\ref{res-vision} shows, on the right image, the detected corners together with the position and orientation estimations of the airplane, based on the left image. The left image is one of the reference images with which image registration is done. The reason for choosing the left image for image registration is because of the best-index variable defined in step two.

\subsection{System Modeling and Control Design}\label{subsec:ctrl}
\begin{figure}[!ht]
\centering
\includegraphics[width=0.6\textwidth]{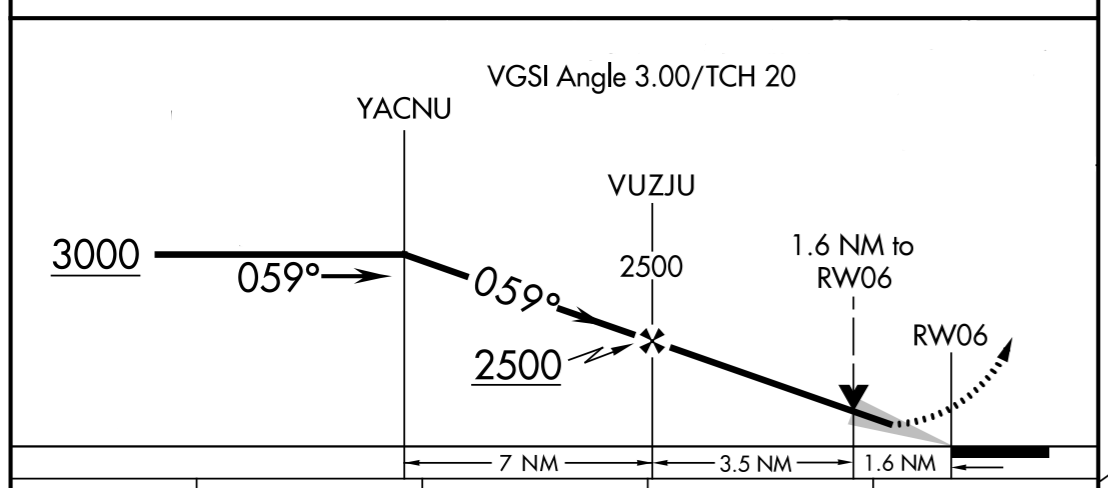}
\caption{Part of the approach chart for Runway 06 at KARB (FAA documentation~\cite{approachChart}). Approach for runway 06 (RW06) goes through waypoints YACNU and VUZJU, has a visual glide slope indicator (VGSI) set to $\mathbf{\alpha}=$3.00${}^\circ$ and a threshold crossing height (TCH) of 20\,ft.
The heading on approach is 059${}^\circ$, the minimum altitudes at YACNU and VUZJU are 3,000\,ft and 2,500\,ft, respectively.
The distance between YACNU and VUZJU is 7\,NM (nautical miles), and the distance between VUZJU and the runway threshold is 3.5+1.6=5.1\,NM.
In this paper we are only looking at the last 800 meters (0.43 nautical miles) of the approach.
Some information has been removed for clarity.
}
\label{fig:approachChartShort}
\end{figure}

\begin{figure}[!ht]
\centering
\includegraphics[width=0.94\textwidth]{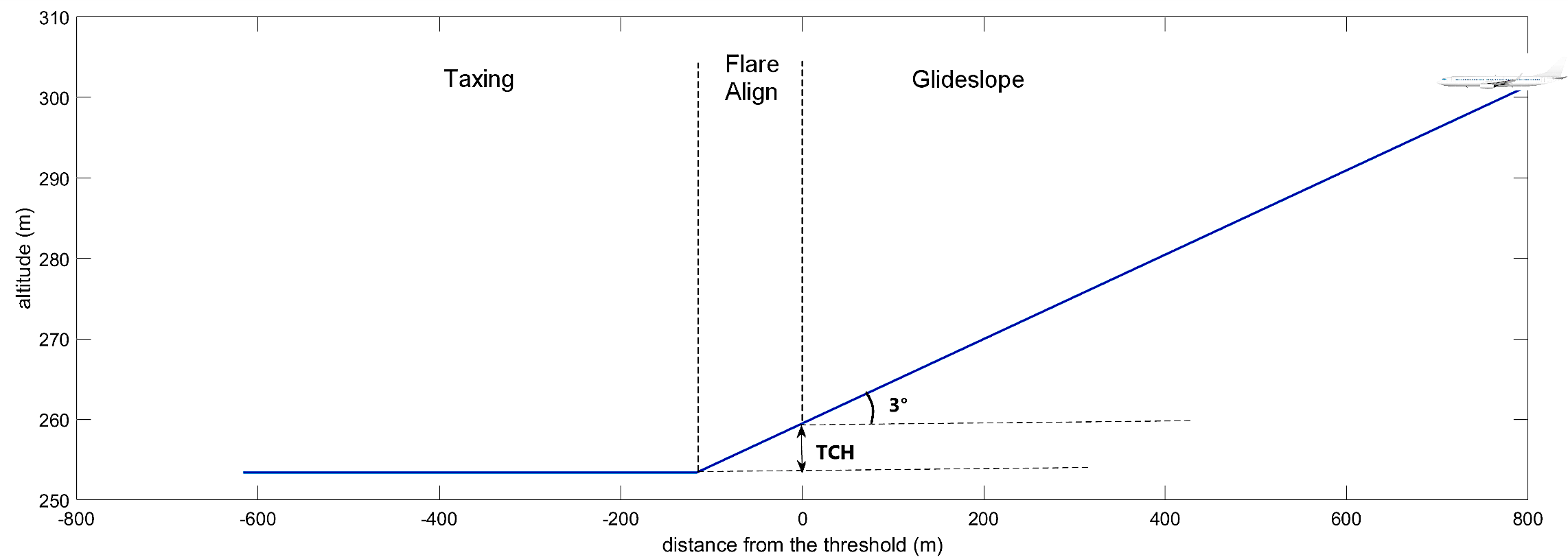}
\caption{Visualization of the generated glideslope for Runway 06 at KARB} 
\label{fig:glideslope}
\end{figure}

Given a specific runway for the final approach, we assume the guidance glideslope to the runway can be defined by an angle and a Threshold
Crossing Height (TCH). This information is provided on the approach charts for a runway.
For Runway 06 at Ann Arbor airport (KARB), the approach plate (Figure~\ref{fig:approachChartShort}) shows 
a typical glideslope angle of 3 degrees, and a TCH of 20\,ft. Using the geometry information for the desired runway, the reference glideslope generator generates several waypoints, forming a consecutive sequence of positions along the correct glideslope from the aircraft's current position to the runway. Figure~\ref{fig:glideslope} shows a sample of glideslope waypoints generated for Ann Arbor airport.

The pre-generated glideslope waypoints are then sent to the closed-loop system to serve as the controller's reference input. To design a proper low-level tracking controller for the automatic landing system, we first identify sufficient variables to represent flight dynamics during the final approach. We use the fixed-wing aircraft dynamics from ~\cite{paw2009synthesis}, where the aircraft is considered a six-degree-of-freedom model described by twelve dynamic or kinematic equations. To briefly present the twelve state variables of the flight dynamics, we consider two reference frames: 1) $O_r X_r Y_r Z_r$ -- the runway threshold frame (a fixed frame with the origin located at the center of the runway threshold and its axes directly backward, left, and vertically upwards with respect to the direction of the runway); and 2) $OXYZ$ -- the aircraft body-fixed reference frame, whose origin is at the center of gravity of the aircraft, with the $X$ axis pointing towards the front of the aircraft, the $Y$ axis pointing to the right side of the aircraft and the $Z$ axis pointing downwards. We then have the following state variables: $V = [u,v,w]^T$, which is the aircraft's linear velocity in the aircraft body frame ($u$ directs towards the aircraft nose, $v$ towards the right wing, and $w$ down)\footnote{We assume there is no wind and therefore the ground speed coincides with the horizontal airspeed.}; the angular velocities $p$, $q$ and $r$ (roll rotation rate, pitch rotation rate, and yaw rotation rate);  the Euler angles $\varphi$, $\theta$, $\psi$ (also expressed in the aircraft frame); $x$, which is the horizontal displacement from the aircraft to the runway threshold; $y$, which is the lateral deviation with respect to the runway; and $h$, which is the flight altitude. All of these state values, which can be obtained or calculated from the combination of the vision data and aircraft instrument data, are then employed by the tracking controller to compute the control signals. Moreover, there are four control inputs, which can change the dynamics of the aircraft: throttle $\delta_t$, elevator deflection $\delta_e$, aileron deflection $\delta_a$, and rudder deflection $\delta_r$. The summary of state variables and control inputs are shown in Table~\ref{tab:sum-cont}.


\begin{table}[!ht]\centering
\caption{\textbf{Summary of state variables and control inputs}}\label{tab:sum-cont}
~\\
\begin{tabular}{lll|lll}
\textbf{Param.} & \textbf{Description} & \textbf{Unit} & \textbf{Param.} & \textbf{Description} & \textbf{Unit}\\
\hline
$u$ & longitudinal velocity & $m/s$&$\varphi$ & roll angle & $deg$\\
$v$ & lateral velocity & $m/s$ &$\theta$ & pitch angle & $deg$\\
$w$ & vertical velocity & $m/s$ & $\psi$ & yaw angle & $deg$\\
$p$ & roll velocity & $deg/s$ &$x$ & horizontal distance & $m$\\
$q$ & pitch velocity & $deg/s$ & $y$ & lateral deviation & $m$\\
$r$ & yaw velocity & $deg/s$& $h$ & aircraft altitude & $m$\\
\hline
$\delta_t$ & throttle control & $N$ &$\delta_r$ & rudder deflection & $deg$\\
$\delta_e$ & elevator deflection & $deg$ & $\delta_a$ & aileron deflection & $deg$\\
\hline
\end{tabular}
\end{table}
The low-level controllers are separately designed for the lateral and longitudinal dynamics. As the aircraft approaches the airport, the lateral dynamics controller aims to decrease the roll angle, yaw angle and lateral deviation with respect to the runway, while the controller for longitudinal dynamics aims to keep a constant descent speed and a three-degree glideslope angle.We implement a proportional-integral (PI) controller for the lateral control, defined as:
\begin{equation}
\begin{split}
&u_{r}(t) = -k_{\psi}^P\psi(t)-k_{y}^Py(t)-k_{\psi}^I\sum_{\tau=0}^{t}\psi(\tau)-k_{y}^I\sum_{\tau=0}^{t}y(\tau),\\
&u_{a}(t) = -k_{\varphi}^P\varphi(t)-k_{\varphi}^I\sum_{\tau=0}^{t}\varphi(\tau).
\end{split}
\end{equation}
where $e_{\psi}$, $e_{y}$, $e_{\varphi}$ are the error states, $k_{\psi}^P$, $k_{y}^P$, $k_{\varphi}^P$ are the proportional coefficients, and $k_{\psi}^I$, $k_{y}^I$, $k_{\varphi}^I$ are the integral coefficients. The output $u_{r}$ and $u_{a}$ of the controller are saturated to compute the input to the system by $\delta_r = sat_{\delta_r^{min}}^{\delta_r^{max}}(u_r)$ and $\delta_a = sat_{\delta_a^{min}}^{\delta_a^{max}}(u_a)$\footnote{Throughout the text, we use the following notation for the saturation function: $sat_{a}^{b}(x)\doteq \max(a,\min(b,x))$.}. Additionally, we implement a PI controller for longitudinal dynamics, which is defined as:
\begin{equation}
\begin{split}
&u_{t}(t) = -k_{u}^P(u(t)-u^{des})-k_{u}^I\sum_{\tau=0}^{t}(u(\tau)-u^{des})+k_t,\\
&u_{e}(t) = -k_{\theta}^P(\theta(t)-\theta_c(t))-k_{q}^Pq(t),\\
&\theta_c(t) = -k_h^P(h(t)-h_c(t))-k_h^I\sum_{\tau=0}^{t}(h(\tau)-h_c(\tau)),\\
&h_c(t) = h_{\rm threshold}+x(t){\rm tan}\gamma.
\end{split}
\end{equation}
where $e_{u}$, $e_{h}$ are the error states, $k_{u}^P$, $k_{\theta}^P$, $k_{q}^P$, $k_{h}^P$ are the proportional coefficients, and $k_{u}^I$, $k_{h}^I$ are the integral coefficients. The output $u_{t}$ and $u_{e}$ of the controller are saturated to compute the input to the system by $\delta_t = sat_{\delta_t^{min}}^{\delta_t^{max}}(u_t)$ and $\delta_e = sat_{\delta_e^{min}}^{\delta_e^{max}}(u_e)$. The parameter values of both the lateral and the longitudinal controller are listed in Tables~\ref{tab:lat_param} and \ref{tab:lon_param} in Appendix~\ref{app:values}.

Figure~\ref{fig:two-cont} presents the path of landing in the vertical and horizontal plane for a Baron 58 (simulated in X-Plane) on Runway 06 at Ann Arbor airport using the designed controller. GPS and other aircraft instrument data have been used for accurate state feedback.

\begin{figure}[!ht]
  \centering
  \begin{minipage}[b]{0.495\textwidth}
    \includegraphics[width=\textwidth]{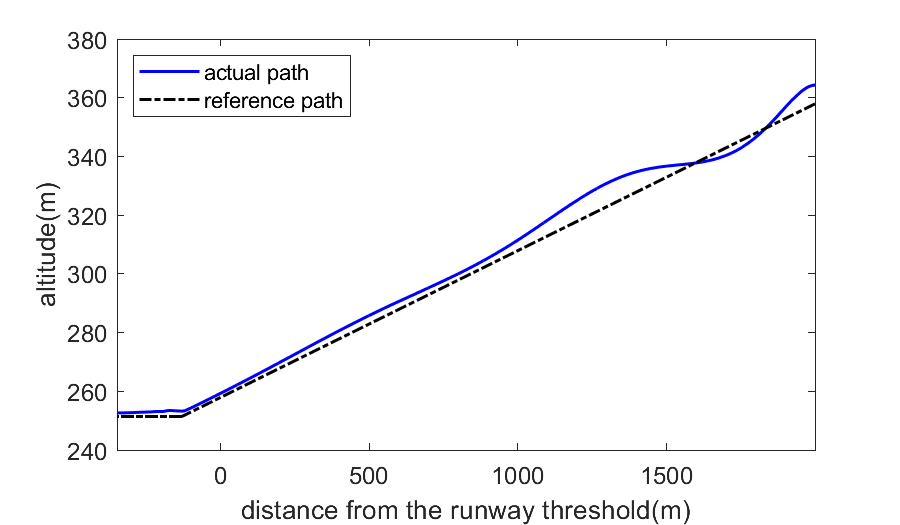}
  \end{minipage}
  \hfill
  \begin{minipage}[b]{0.495\textwidth}
\includegraphics[width=\textwidth]{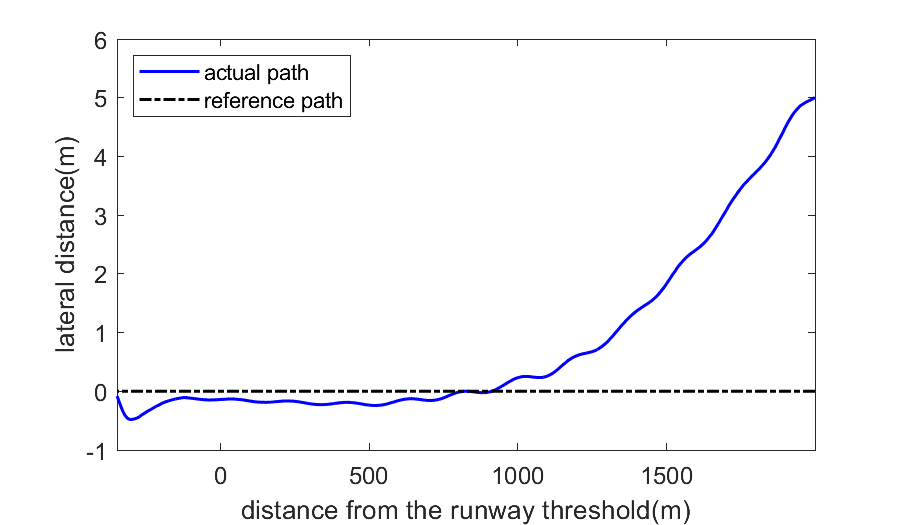}
  \end{minipage}
  \caption{Path of landing for Baron 58 in the vertical plane (left) and the horizontal plane (right)} 
  \label{fig:two-cont}
\end{figure}



\subsection{System integration}
\begin{figure}[!ht]
\centering
\includegraphics[width=0.99\textwidth]{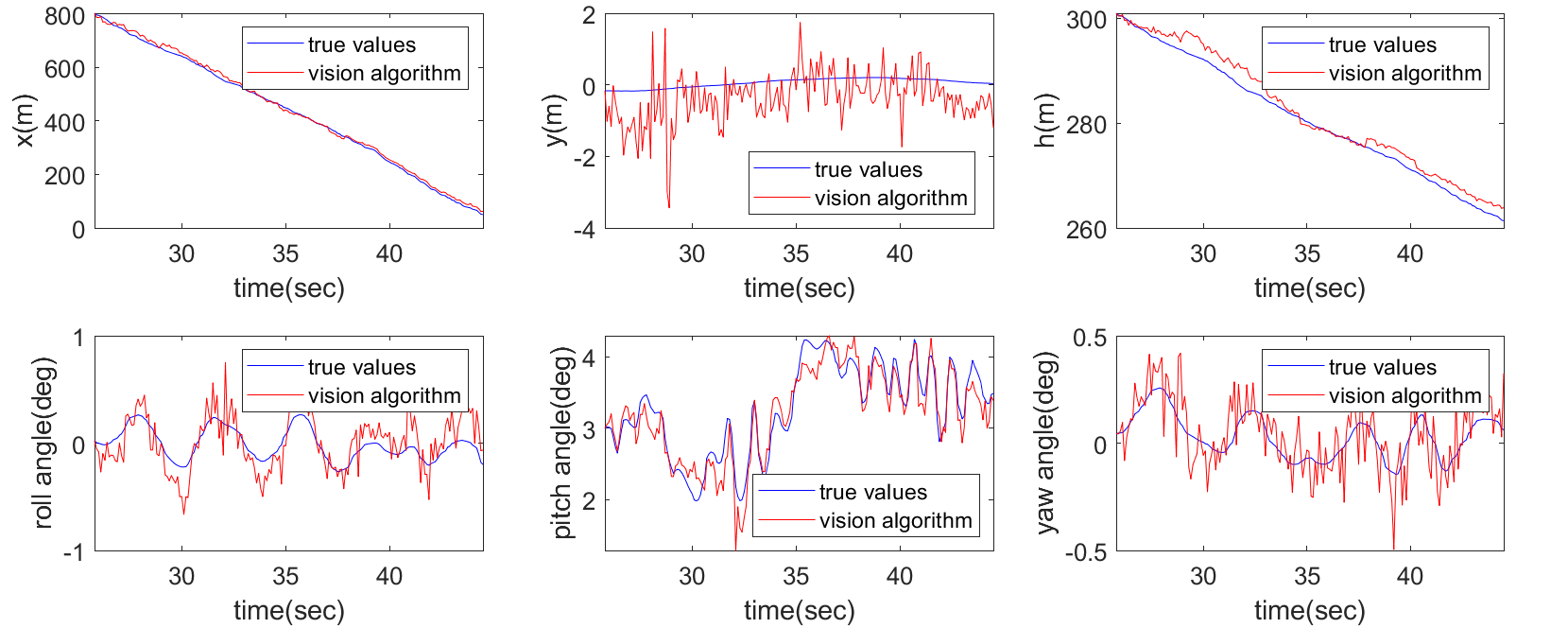}
\caption{The comparison between the values taken from vision-based estimator and ground-truth.}
\label{fig:landingHfromVision}
\end{figure}

In this section, we integrate the controller described in section~\ref{subsec:ctrl} with the vision-based estimator described in section~\ref{subsec:vision}. The overall system is implemented in MATLAB/Simulink by integrating several modules. We use X-Plane as the flight dynamics simulator and use the X-Plane Connect Toolbox~\cite{xplaneconnect} for interfacing between the controllers, vision modules, and X-Plane.
The values of $x$, $y$, $h$, $\theta$, $\varphi$, and $\psi$ in state feedback are computed through the vision pipeline, and the values of $u$, $q$ are taken from the instrument data in X-plane. Figure~\ref{fig:landingHfromVision} shows the comparison between the state values from the vision-based estimator and the ground-truth during the landing approach. Although the estimation from the vision-based estimator is noisy, the error between the estimation and the ground-truth values are close to tolerable ranges, discussed further in Section~\ref{sec:falsif_analysis}. In particular, the estimator is able to catch small changes of the orientation.

\begin{figure}[!ht]
\centering
\includegraphics[width=0.995\textwidth]{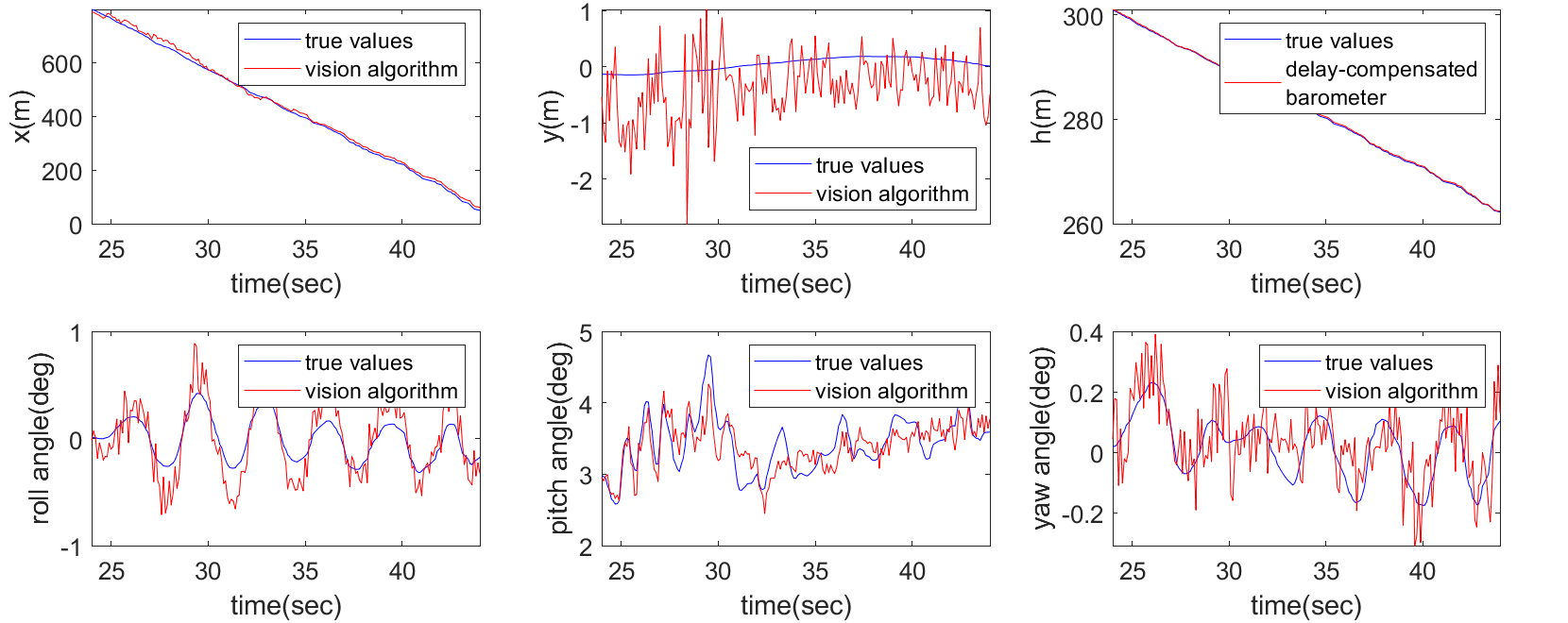}
\caption{The comparison between values in state feedback and ground-truth when landing with the vision-based estimator with $h$ from the delay-compensated barometer}
\label{fig:landing-IPPE-ASPnP}
\end{figure}

Since the accuracy of $h$ is crucial during the landing approach, we also implemented a method to get the value of $h$ from the barometer and feed it to the vision algorithm to enhance the estimation of the orientation, aircraft's altitude, and lateral deviation. Moreover, a Kalman filter is implemented to reduce the delay of the barometer. Figure~\ref{fig:landing-IPPE-ASPnP} shows the performance of the vision-based estimator during the landing by feeding the value of $h$ taken from the delay-compensated barometer. The mean and the standard deviations of the  absolute error values between estimated variables and the ground-truth are shown in Table~\ref{tab:false-ex-num}. 
The error from the estimation of $h$ is reduced when using the delay-compensated barometer while the errors from other states remain similar. 


\begin{table}[!ht]\centering
\caption{\textbf{Mean and the standard deviations of the  absolute error values between estimated variables and the ground-truth. $h_{baro}$ indicates the height data taken from the delay-compensated barometer.}}\label{tab:false-ex-num}
\renewcommand{\arraystretch}{1.2}
\begin{tabular}{|c|c|c|c|c|c|c|c|}
\cline{3-8}
\multicolumn{1}{c}{}&\multicolumn{1}{c|}{}&$x(m)$&$y(m)$&$h(m)$&$\theta(deg)$&$\varphi(deg)$&$\psi(deg)$\\ \hline
\multirow{2}{*}{\shortstack{landing with\\vision-based estimator}}&mean&$4.4422$&$0.2768$&$0.6851$& $0.0577$& $0.0691$& $0.0425$\\ \cline{2-8}
&\shortstack{standard\\deviation}& $6.2414$&$0.4245$ &$1.0671$ &$ 0.0887$ & $0.0156$& $0.0647$ \\ \hline
\multicolumn{1}{c}{}&\multicolumn{1}{c|}{}&$x(m)$&$y(m)$& $h_{baro}(m)$&$\theta(deg)$&$\varphi(deg)$&$\psi(deg)$\\ \hline
\multirow{2}{*}{\shortstack{landing with $h_{baro}$ and\\ vision-based estimator}}& mean &$4.4674$& $0.2484$ &$0.0329$& $0.0705$& $0.1038$& $0.0376$\\ \cline{2-8}
&\shortstack{standard\\deviation} &$6.2002$ &$0.3791$ &$0.0415$&$0.1063$ & $0.1530$& $0.0556$  \\ \hline
  \end{tabular}
\end{table}

\section{Falsification-driven analysis of the vision-based automatic landing system}

In order to evaluate the proposed automatic landing architecture, the developed landing controller, and perception-based estimation modules, we use the falsification tool Breach~\cite{donze2010breach}. To be able to apply falsification, we first formalize the specifications for landing.

\subsection{Formal specification of the landing system}
Landing specifications include (i) a bound on the maximal deviation from the  reference glideslope, (ii) bounds on the maximal lateral and longitudinal deviation from the target landing point to avoid missing the runway, and (iii) bounds on the airspeed and vertical velocity in a period slightly before and after touchdown to ensure smoothness of the landing and to avoid potential damage to the landing gear. Landing can be separated into three phases: (i) final approach, (ii) flare, and (iii) touchdown. In this paper, we focus on the final approach and formalize final approach specifications using linear temporal logic~\cite{pnueli1977temporal}. 


 Final approach, which is the first phase of landing, usually begins with either a turn which aligns the plane with the runway or at an initial approach point, and ends when the altitude is less than 15 ft. Specifications for the approach are based on FAA sources and pilot experience. They are intended to ensure that the airplane remains safe by controlling speeds, pitch, roll, and yaw based on recommendations for a stabilized approach and landing. The specifications do not account for weather, including crosswind yaw corrections and any headwinds or tailwinds, but this eventually can be added, and the bounds of the specifications allow for some corrections due to wind. An angle of attack specification was not included because stalls can reasonably be prevented through controlling airspeed and pitch, and angle of attack data is not easily accessible in most general aviation aircraft or obtainable through the vision algorithm.  The first three specifications are related velocity in $x$, $y$, and $z$ direction, and the last two are lateral deviations from the runway and glide path~\cite{federal2011airplane,MARSH:2020,saini2010flying,AviationChief:2018}. A visual summary of the specifications from both a top and side view can be found in Figure~\ref{fig:specTop}.
The definitions of each specification are explained below.

\subsection*{Specification on longitudinal speed: $\Phi_1  \equiv u_c V_{so} - u_l \leq u \leq u_c V_{so} + u_u$}
%
%
%
\noindent
The variable $u$ is the velocity in the $x$ direction. In order to maintain a stabilized approach or a constant angle glidepath without the need for significant pilot adjustment, $u$ should be close to $u_cV_{so}$, where typically $u_c=1.3$. $V_{so}$ is the stall speed of the airplane. Stall speed is different for each plane depending on the shape of the wing and other aerodynamic factors, but it represents the speed where the plane loses lift and begins to stop flying. There are bounds on either side of the specification to allow for small deviations. The airplane should not be much slower than $u_cV_{so}$ on landing but can be slightly faster, so the bounds are uneven. 

\subsection*{Specification on lateral speed: $\Phi_2 \equiv -\delta_v  \leq v \leq \delta_v $} 
\noindent
The variable $v$ is the velocity in the $y$ direction. In order to stay close to the axis of the runway during the entire final approach, the lateral velocity of the airplane should be close to zero. The bounds are defined to allow for corrections needed due to changes in wind or small misalignment so that the pilot can get the plane back on the extended center line of the runway, which would require a small lateral velocity.  

\subsection*{Specification on vertical speed: $\Phi_3 \equiv w_l  \leq w \leq w_u \tan(\alpha) u $}
\noindent
The variable $w$ represents the velocity in the $h$ direction or climb/descent rate. Since the Z axis points downward, $w$ is positive during approach or when the airplane descends. A constant angle glideslope with the ratio of $3:1$ should be maintained for a stabilized approach, which means for every $3$ nautical miles (nm) or around 18228 ft flown over the ground, the aircraft should descend 1000 ft (about $3$ degrees). The specification does not allow to climb during the final approach as the descent should be constant, and adjustments to a large rate of descent can be made by maintaining altitude as opposed to climbing. The specification allows $w$ to be slightly larger than the required rate  to account for re-adjustments due to wind or pilot error; however, $w$ must still be within the boundary of $w_u$ times the standard rate of descent for a 3-degree glideslope. 

\subsection*{Specification on lateral deviation: $\Phi_4 \equiv -(x + d_r)\tan({\beta}) \leq y \leq (x + d_r)\tan({\beta}) $}
\noindent
The variable $y$ is the lateral deviation from the runway. Similarly to specification on lateral speed $\Phi_2$, the airplane should stay close to the extended center line of the runway during the final approach. Small deviations up to the bound are allowed to correct for wind or misalignment due to pilot error. The distance that the airplane needs to be within from the center of the runway is determined by a deviation angle as opposed to a numerical distance, so the airplane needs to be closer to the runway center line  as the closer it gets to the touchdown point, as shown in Figure \ref{fig:specTop}a. 

\subsection*{Specification on vertical position: $\Phi_5 \equiv (x + d - t)\tan{(\alpha - \alpha_h}) \leq h \leq (x + d + t)\tan{(\alpha + \alpha_h)}$}
\begin{figure}[h]

\begin{subfigure}[b]{0.5\textwidth}
\includegraphics[width=\textwidth]{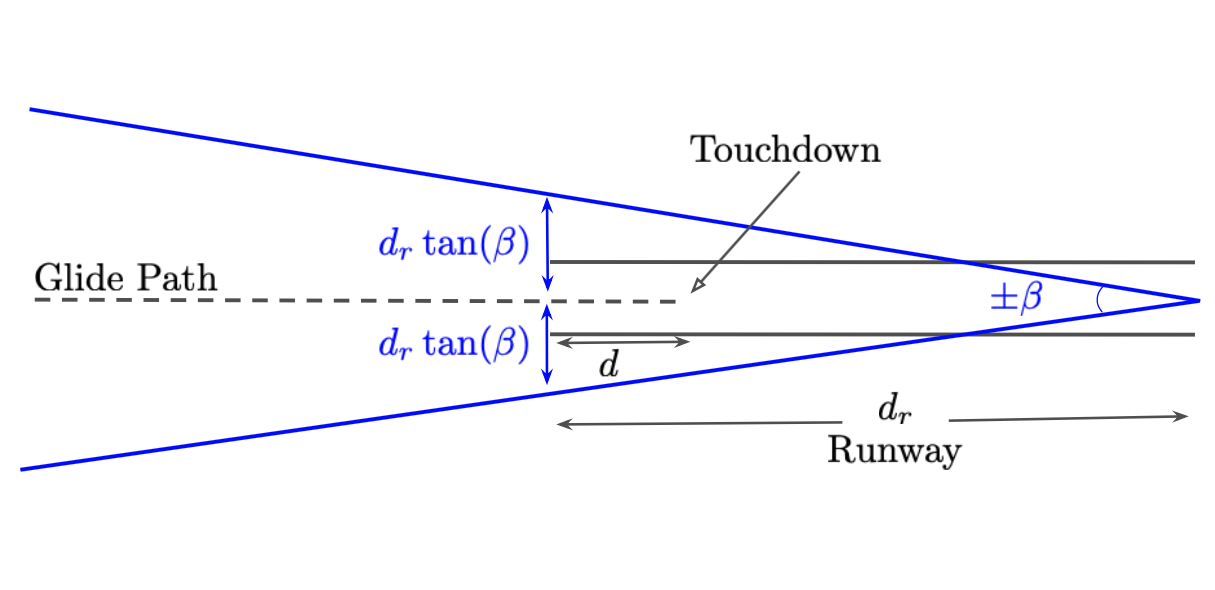}
\caption{Top view}
\end{subfigure}
\begin{subfigure}[b]{0.5\textwidth}
\includegraphics[width=\textwidth]{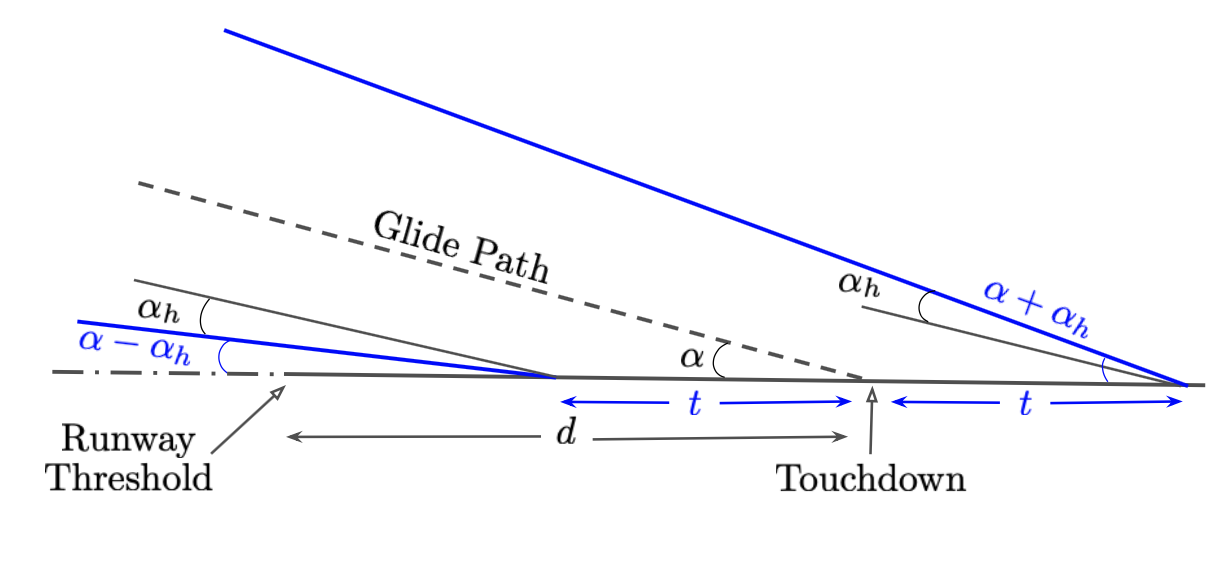}
\caption{Side view}
\end{subfigure}
\caption{Visual representation of specifications for landing (Top View on the left and and Side view on the right). The blue lines represent the specifications.}
\label{fig:specTop}
\end{figure}

\noindent
The variable $h$ is the height above the airport. 
The specification indicates that an airplane should stay within a small offset of the glideslope during the final approach. The angle of the glideslope has a bound on either side which corresponds approximately to the glideslope angle where a pilot would see either three red or three white lights on a PAPI (Precision Approach Path Indicator), and therefore still be within a safe distance from the glideslope. The visual representation of the specification is shown in Figure \ref{fig:specTop}b.
\\\\
Putting all the specifications together, the LTL specification for final approach is given by: 
\begin{equation}\label{eq:spec_all}
\Phi  \equiv  \left(\bigwedge_{i=1}^5  \Phi_i\right)\  \mathcal{U}\  (h \leq h_f \ft),
\end{equation}m
where $h_f$ is the altitude where final approach ends and flare phase starts.  Table~\ref{tab:spec_vars} shows the different parameters used in our experiments.

\begin{table}[!ht]\centering
\caption{The variables used in each specification and typical numerical values in SI and aviation standard units. }\label{tab:spec_vars}
\renewcommand{\arraystretch}{1}
\begin{tabular}{|m{1.5cm}<{\centering}|m{8cm}<{}|m{2cm}<{\centering}|m{2.5cm}<{\centering}|}
\hline
{\bf Variable} & \centering {\bf Description}& {\bf Value in SI} &
{\bf \shortstack{Value in Aviation\\Standard Units}}\\
\hline
$u_c$ & $x$-velocity constant & $1.3$ & $1.3$\\  
\hline
$u_l$ & Lower bound of velocity in $x$ direction & $2.6 \ms$ &$5 \kts$ \\
\hline
$u_u$ & Upper bound of velocity in $x$ direction & $5.1 \ms$ &$10 \kts$\\
\hline
$\delta_v$ & Bound of velocity in $y$ direction & $1.51 \ms$ & $3 \kt$\\ 
\hline
$\alpha$ & Glide slope angle & $3^{\circ}$ & $3^{\circ}$\\  
\hline
$w_l$ & Lower bound of velocity in $z$ direction & $0 \ms$  &  $0 \fpm$ \\
\hline
$w_u$ & Upper bound of velocity in $z$ direction & $2$ &  $2$\\  
\hline
$d_r$ & Distance to the end of the runway (Intersection) & $3048 \m$ &  $10000 \ft$\\  
\hline
$\beta$ &Lateral Deviation Angle & $2^{\circ}$ &  $2^{\circ}$\\  
\hline
$d$ & Distance from the runway threshold to touch down marks & $305 \m$ &  $1000 \ft$\\  
\hline
$t$ & Touchdown range & $305 \m$ &  $1000 \ft$\\  
\hline
$\alpha_h$ & Glideslope angle bound & $0.7^{\circ}$ &  $0.7^{\circ}$\\ 
\hline
$h_f$ & Altitude at end of final approach & $5 \m$ &  $15 \ft$\\  
\hline

\end{tabular}
\end{table}

\subsection*{Validation with actual flight data}
\begin{figure}[!h]
\centering
\includegraphics[width=1\textwidth]{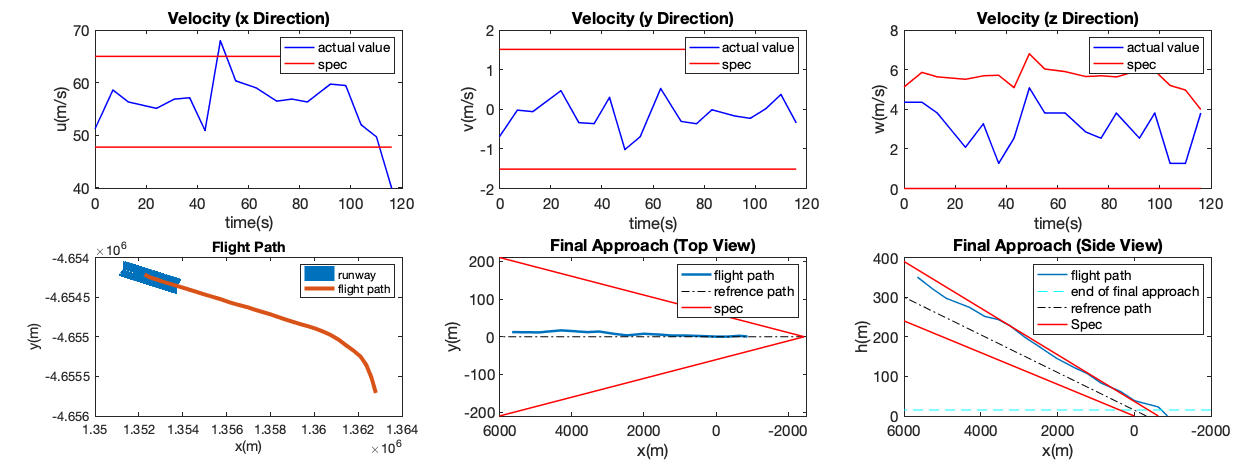}
\caption{The graphs are for a flight from Washington (IAD) to New York (JFK). The airplane used in this flight is a Mitsubishi CRJ-900LR. The flight duration was 42 minutes}
\label{fig:dataJFK}
\end{figure}

\begin{figure}[!h]
\centering
\includegraphics[width=1\textwidth]{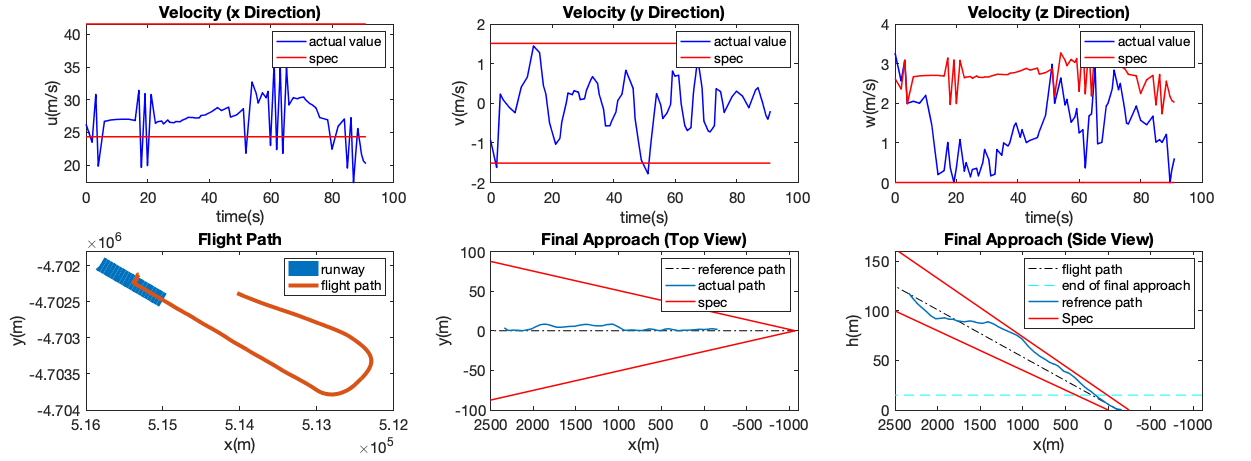}
\caption{Data is from a flight by a private pilot (Isabelle Gallagher, an author of this paper) from a flight over Ann Arbor airport in a Cessna 172. It is important to note that the ideal touchdown point is different for the two airports because there are different TCH's for different airports. TCH is the minimum height that the airplane needs to be above the ground when it reaches the runway threshold \cite{tch}. It is 54ft for JFK~\cite{JFKRunway} which puts the ideal touchdown point at 1030ft into the runway, and 20ft for Ann Arbor airport~\cite{approachChart} which puts the ideal touchdown point at 400ft into the runway  
}
\label{fig:dataARB}
\end{figure}

To ensure that our specifications represent a typical final approach well, we have plotted the data from two different flights to determine whether they satisfy the defined specifications. The first set of graphs, which is shown in Figure~\ref{fig:dataJFK}, are the data taken from a commercial flight tracking website~\cite{fr24} for a flight in a CRJ-900LR airplane. For this dataset, all the specifications for the position of the airplane are satisfied except for the height at the very end of the approach. The specifications for vertical and lateral velocity are also satisfied. However, specifications for the longitudinal velocity of the aircraft were violated at one point. The second set of graphs, shown in Figure~\ref{fig:dataARB} are the data taken from a flight over Ann Arbor in a Cessna 172. The graphs show that the specifications for the aircraft's position are all satisfied along with the specification for lateral and vertical velocity, while the specifications for longitudinal velocity are not. We believe there could be noise caused by imperfect measurements from the sensors. This could be why some specifications are not satisfied. Also, the exact numbers used for specifications are rough estimates, and pilots do not have to follow them exactly. 



\subsection{Falsification-based analysis} \label{sec:falsif_analysis}
We use the Breach falsification tool  \cite{donze2010breach}, which can search for initial conditions, parameter values, disturbance, and noise sequences within the given bounds to describe the system's environment to try to locate trajectories that violate a given specification. 
In the following, we discuss several falsification results of the system with and without the vision-based estimator. All of the examples are tested without wind and we only focus on the performance during the final approach. Because the velocities of aircraft in X-plane cannot be initialized at arbitrary values, at the start of each simulation, the aircraft is accelerated on the runway to get enough longitudinal velocities for the final approach and then placed at a location on the glideslope -- but not at the starting point of the final approach. For landing a Baron 58 onto Runway 06 at KARB, we place the aircraft with $x=2000\m$ on the glideslope. The aircraft is then controlled to move along the glideslope. The falsifier only checks the performance or adds disturbances when $x<800\m$, which is the normal distance of the final approach for Runway 06 at KARB. The landing process between $800\m \leq x \leq 2000\m$ is used for the aircraft to obtain appropriate vertical and lateral velocities for the final approach. For vision-in-the-loop, the vision-based estimator is involved when $x<800\m$. In the following section, we denote $x_{\rm far}=2000\m$ and $x_0=800\m$.

First, we conduct falsification of the system without the vision-based estimator. Under these conditions, all the values used in the state feedback are taken from the X-Plane simulation engine, which indicates the ground-truth positions and velocities of the aircraft in X-Plane. We consider each state individually and increment its allowable noise in the Breach falsifier until the specification is falsified. Table~\ref{tab:false-ex} shows the maximum tolerable bounds of noises for which Breach cannot falsify the specification. Once we have these maximum tolerable noise bounds, we can compare them to the vision system's estimation error. More specifically, at each cycle of falsification, the Breach tool only adds noise to one state, and it is limited to five samples for each bound. The tolerable noise bounds indicate that if Breach is allowed to choose noises larger than that bound, falsifying trajectory can be generated within five samples.

\begin{table}[!ht]\centering
\caption{\textbf{The tolerable bound of noises of each state, $h_{td}$ indicates the altitude at the touch down point, which is 253m at the KARB}}\label{tab:false-ex}
\renewcommand{\arraystretch}{1.2}
\begin{tabular}{|m{2cm}<{\centering}|m{6cm}<{\centering}|m{1cm}<{\centering}|m{3.5cm}<{\centering}|}
\hline
\bf{noisy state}& \bf{tolerable bound of noises}&\bf{unit} &\bf{\shortstack{potentially falsified\\ specifications}}\\
\hline
$u$ & $[-4.5,4.5]$ & $m/s$ & $\Phi_1$\\  
\hline
$y$ & $[-6.5,6.5]$ & $m$ &$\Phi_2$ \\ 
\hline
$\varphi$ & $[-2,2]$ & $deg$ &$\Phi_2$ \\  
\hline
$\psi$ &$[-2.5,2.5]$ & $deg$ & $\Phi_2$\\ 
\hline
$x$ & $[-15\%x,15\%x]$ & $m$ & $\Phi_3$ \\  
\hline
$h$ & $[-16\%(h-h_{td}),16\%(h-h_{td})]$ & $m$  &  $\Phi_3$ \\
\hline
$\theta$ & $[-6,6]$ & $deg$ &  $\Phi_3$\\  
\hline
$q$ & large than its magnitude & $deg/s$ &  $\Phi_3$\\  
\hline
\end{tabular}
\end{table}

\begin{figure}[!ht]%
 \centering
 \subfloat{\includegraphics[width=0.87\textwidth]{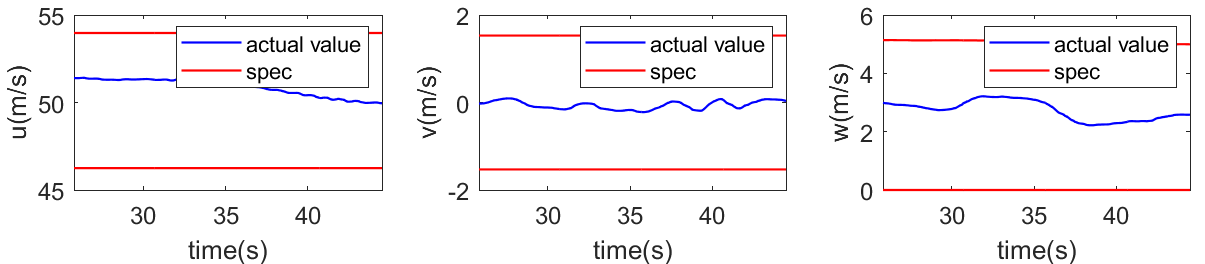}}\\
 \subfloat{\includegraphics[width=0.49\textwidth]{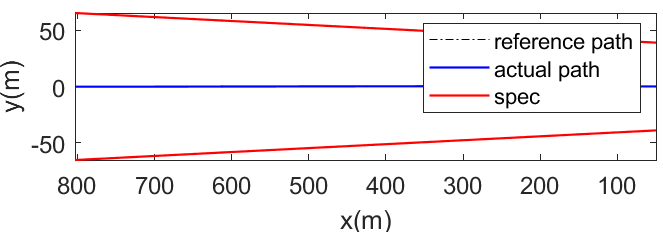}}
 \subfloat{\includegraphics[width=0.49\textwidth]{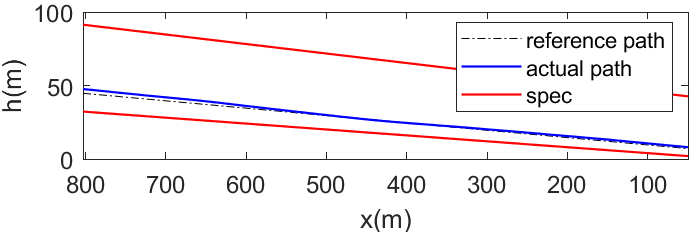}}%
 \caption{The specifications and actual values when landing with the vision-based estimator}%
 \label{spec-vision}%
\end{figure}

\begin{figure}[!ht]%
 \centering
 \subfloat{\includegraphics[width=0.9\textwidth]{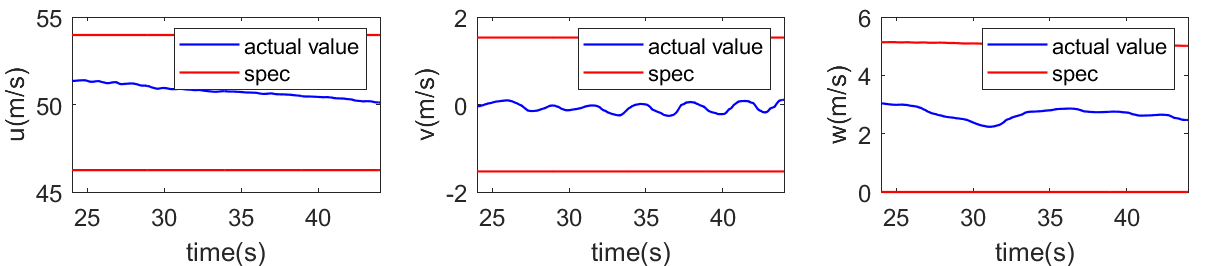}}\\
 \subfloat{\includegraphics[width=0.492\textwidth]{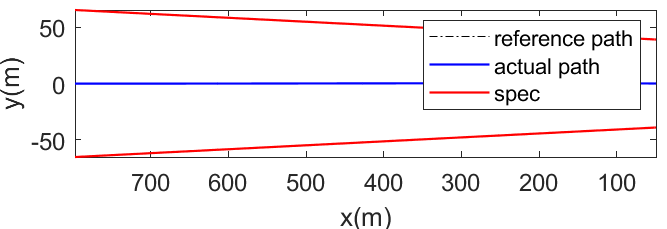}}
 \subfloat{\includegraphics[width=0.492\textwidth]{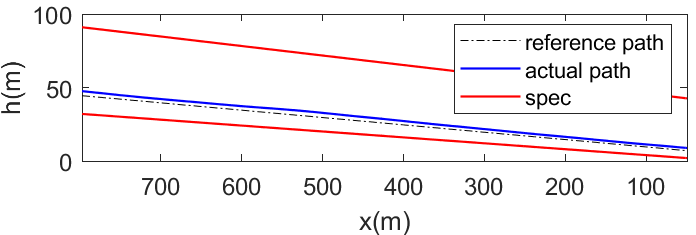}}%
 \caption{The specifications and actual values when landing with the vision-based estimator with $h$ from modified barometer}%
 \label{spec-IPPE-ASPnP}%
\end{figure}

For vision-in-the-loop, the values of $x$, $y$, $h$, $\theta$, $\varphi$, and $\psi$, which are used in the state feedback, come from the vision-based estimator or the barometer. Therefore, if errors between values from the vision-based estimator or the barometer and ground-truth are larger than tolerable bounds, the specifications will likely fail when vision is in the loop, according to the Breach falsification results of the system without the vision-based estimator. Two different cases are tested. First, landing with the vision algorithm in the loop is tested. All six states are taken from the vision-based estimator. Second, the variable $h$ is replaced by the height data from the modified barometer. Because all of the errors are smaller than the tolerable bounds under these two circumstances, none of the specifications are falsified, as shown in Figure~\ref{spec-vision} and Figure~\ref{spec-IPPE-ASPnP}. 

In the previous falsification examples, we focus on the trajectories starting on the glideslope with and without the vision-based estimator. Here we analyze how initial deviations from the glideslope affect specification satisfaction using ground-truth positions and comparing the result with the vision-in-the-loop. 

\begin{figure}[!ht]
\centering
   \begin{subfigure}{0.48\linewidth} \centering
     \includegraphics[width=0.8\textwidth]{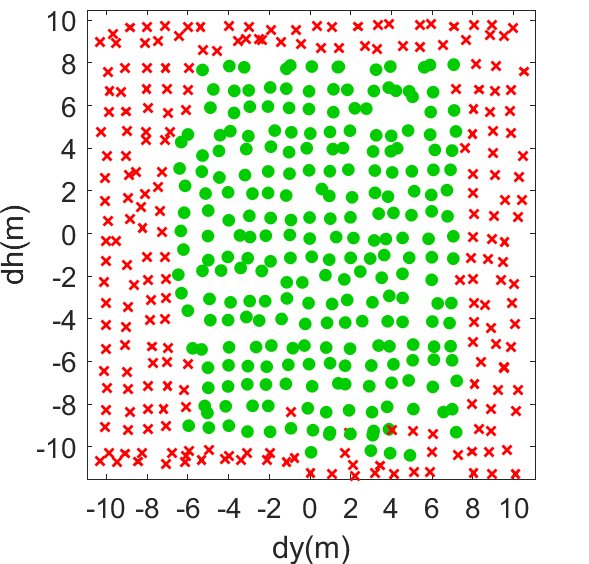}
     \caption{}
     \label{fig:initial-noVision-3kt}
   \end{subfigure}
   \begin{subfigure}{0.48\linewidth} \centering
     \includegraphics[width=0.8\textwidth]{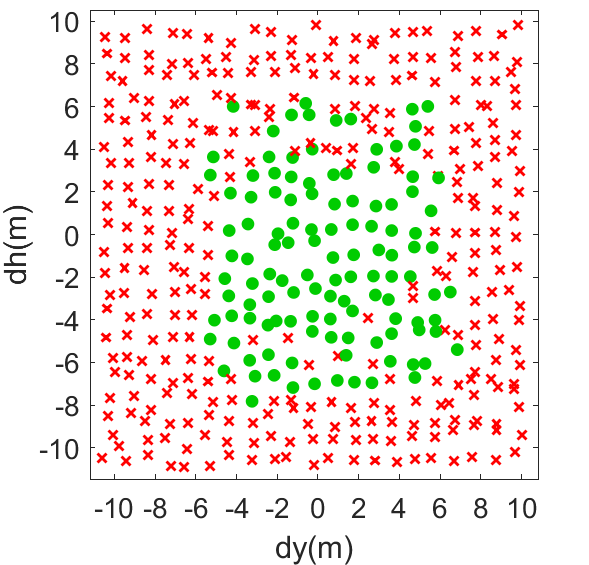}
     \caption{}
     \label{fig:initial-vision-3kt}
   \end{subfigure}
\caption{(a) Satisfaction distribution plot of system without the vision-based estimator. (b) Satisfaction distribution plot of vision-in-the-loop.}
\label{fig:initial}
\end{figure}

We start the simulation with deviations in the direction of $h$ and $y$. The initial set of deviations are $dy\in[-10,10]$, and $dh\in[-10,10]$ (in the unit of the meter) and are meshed into a $21\times 21$ grid. In X-plane, the aircraft needs to have the landing process between $x_0 \leq x \leq x_{\rm far}$ to obtain the required vertical velocity and the lateral velocity. During this process, the aircraft is controlled along the glideslope with deviations. When $x<x_0$, the aircraft is controlled to move along the original glideslope. However, because of the controller's limits, the deviations at $x_0$ may not be the same as the deviations at $x_{\rm far}$. Figure~\ref{fig:initial} is a set of irregular plots, meaning that the points in the plots show the actual deviations from the glideslope at $x_0$. Red points indicate  the falsifying initial conditions; green dots indicate the non-falsifying initial conditions. 


Figure~\ref{fig:initial-noVision-3kt} shows that the deviations between $dy\in[-6,7]$ and $dh\in[-9,8]$ are acceptable for the system without the vision-based estimator. Some asymmetrical features of the aircraft may cause the asymmetry between the lower bound and the upper bound of $dy$ and $dh$, but this is not fully understood. Figure~\ref{fig:initial-vision-3kt} shows that the specifications are not falsified when $dy\in[-4,6]$, $dh\in[-7,6]$ for vision-in-the-loop. Because the errors that come from the vision-based estimator can add other disturbances to the system, the acceptable deviations become smaller than the system without the vision-based estimator. Additionally, since the vision-based estimator's errors are not systematic and somewhat unpredictable, this leads to the blurry boundaries between the falsifying and non-falsifying initial conditions. 

\begin{figure}[!ht]
\centering
    \includegraphics[width=0.95\textwidth]{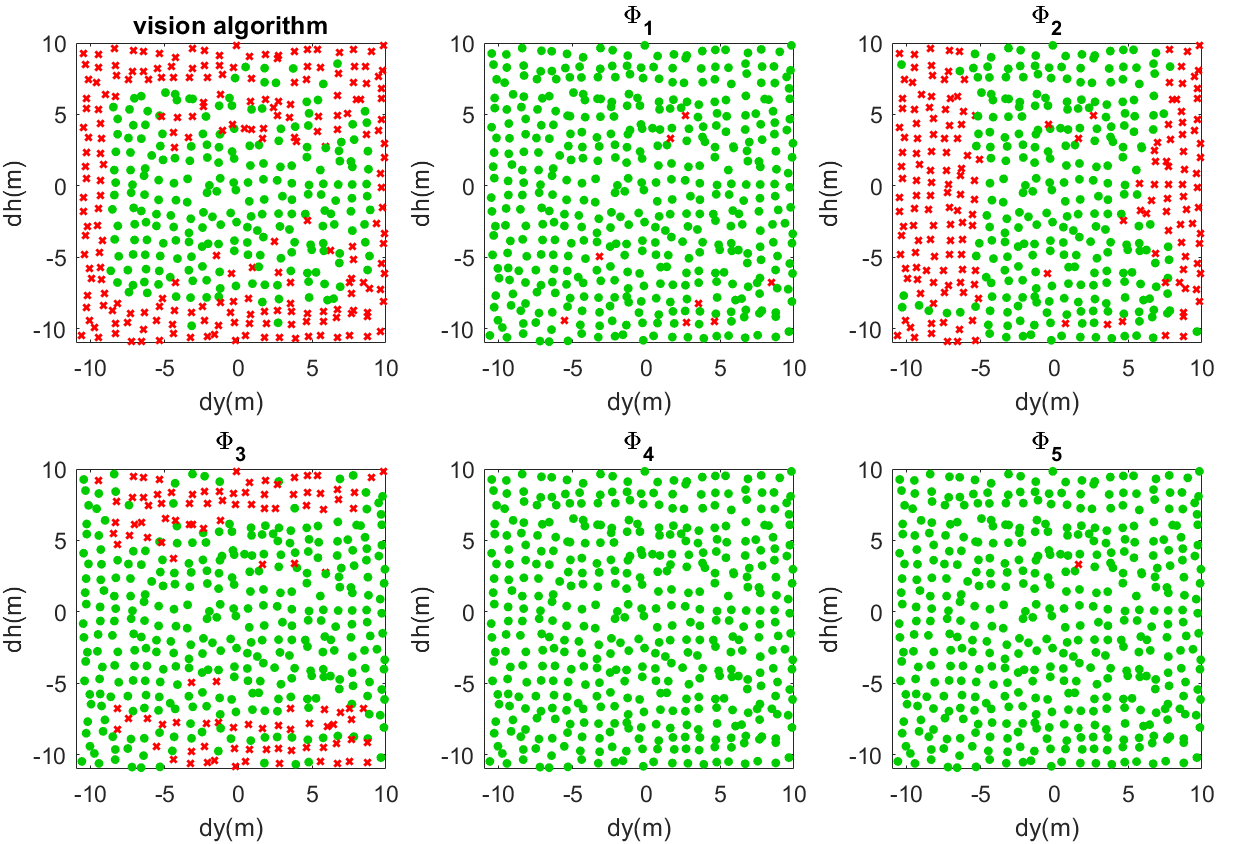}
    \caption{Satisfaction distribution plot of each specification. The falsifying initial conditions in the top left plot indicate that the vision algorithm fails to provide reasonable outputs.}
\label{fig:spec_performance}
\end{figure}

Figure~\ref{fig:spec_performance} shows the satisfaction distribution of each specification when landing with the vision-based estimator under different initial conditions. Red crosses and green dots share the same meaning with those in Figure~\ref{fig:initial}. The figure indicates that $\Phi_2$ tends to be falsified when the initial conditions have large lateral deviations, and $\Phi_3$ tends to be falsified when the initial conditions have large vertical deviations. $\Phi_1$, $\Phi_4$, and $\Phi_5$ are loose enough for the final approach so they are not falsified under most of the situations. Moreover, the falsifying initial conditions in the top left subplot indicate that the vision-based estimator fails to provide reasonable outputs during the landing approach, which leads to the failure of landing. Usually, the failure occurs when the aircraft oscillates severely and the runway disappears in the image fed into the vision-based estimator. 

\section{Conclusions and future work}
 This paper presented a prototype automatic landing system and a framework for analyzing such systems via falsification. We described three significant parts of this system: (i) the image processing algorithms that can identify the runway from images captured from the X-Plane flight simulator and estimate the distance and orientation information; (ii) the reference glideslope generator which can generate waypoints for the desired glidepath; and (iii) the low-level controllers that keep the aircraft on the desired glideslope.  The paper can be divided into two parts: (i) integrating the vision-based perception with the automatic landing system to serve as a supplement to the instrument data; and (ii) formalizing the landing specifications into requirements presented by signal temporal logic, then employing various falsification techniques to evaluate both tracking controllers and image processing algorithms to find any counterexamples which do not satisfy the required specifications. 

For the future work, 
monitors (similar to monitors for run-time verification \cite{bartocci2019automatic}) can be used to compare the results of vision-based position estimation (i.e., the perception module) with the ``actual'' position information from X-Plane. 
 Then, these monitored values can be used to provide diagnostic information when a counterexample for the system-level specification is found, and to try to understand the landscape of root-causes of failures found by the falsification tool. Additionally, the spread of identified failures in different parts of the architecture can be identified: are failures mostly due to an error in perception, an error in the controller, or both? 



\paragraph{Acknowledgments:} This work is supported in part by Collins Aerospace,  a unit of Raytheon Technologies. The authors would like to thank Eelco Scholte and Claudio Pinello from Collins Aerospace and the United Technologies Research Center (UTRC) for helpful discussions. The article solely reflects the opinions and conclusions of its authors and not Collins Aerospace.

\bibliography{ref}
\newpage
\appendix
\section{Parameter values for controllers}
\label{app:values}
\begin{table}[htbp]\centering
\caption{\textbf{Parameter values for the lateral controller}}\label{tab:lat_param}
\begin{tabular}{lll}
\textbf{Param.} & \textbf{Description} & \textbf{Value}\\
\hline
$k_{\psi}^P$ & proportional coefficient for $\psi$ & $1 \ (deg/deg)$\\
$k_{y}^P$ & proportional coefficient for $y$ & $0.5 \ (deg/m)$\\
$k_{\varphi}^P$ & proportional coefficient for $\varphi$ & $1 \ (deg/deg)$\\
$k_{\psi}^I$ & integral coefficient for $\psi$ & $0.1 \ (deg/deg)$\\
$k_{y}^I$ & integral coefficient for $y$ & $0.01 \ (deg/m)$\\
$k_{\varphi}^I$ & integral coefficient for $\varphi$ & $0.1 \ (deg/deg)$\\
$\delta_r^{min}$ & minimum rudder deflection angle & $-27 \ (deg)$\\
$\delta_r^{max}$ & maximum rudder deflection angle & $27 \ (deg)$\\
$\delta_a^{min}$ & minimum aileron deflection angle & $-20 \ (deg)$\\
$\delta_a^{max}$ & maximum aileron deflection angle & $20 \ (deg)$\\
\hline
\end{tabular}
\end{table}
\begin{table}[htbp]\centering
\caption{\textbf{Parameter values for the longitudinal controller}}\label{tab:lon_param}
~\\
\begin{tabular}{lll}
\textbf{Param.} & \textbf{Description} & \textbf{Value}\\
\hline
$\upsilon^{des}$ & desired velocity in the direction of $\upsilon$ & $50 \ (m/s)$\\
$k_{t}$ & thrust constant & $5000 \ (N)$\\
$h_{\rm threshold}$ & desired altitude at the threshold & $259.51 \ (m)$\\
$\gamma$ & glideslope angle & $3 \ (deg)$\\
$k_{\upsilon}^P$ & proportional coefficient for $\upsilon$ & $70 \ (N/m/s)$\\
$k_{\theta}^P$ & proportional coefficient for $\theta$ & $8 \ (deg/deg)$\\
$k_{q}^P$ & proportional coefficient for $q$ & $0.05 \ (deg/deg/s)$\\
$k_{h}^P$ & integral coefficient for $h$ & $0.3 \ (deg/m)$\\
$k_{\upsilon}^I$ & integral coefficient for $\upsilon$ & $2 \ (N/m)$\\
$k_{h}^I$ & integral coefficient for $h$ & $0.01 \ (deg/m)$\\
$\delta_t^{min}$ & minimum engine thrust & $0 \ (N)$\\
$\delta_t^{max}$ & maximum engine thrust & $10000 \ (N)$\\
$\delta_e^{min}$ & minimum elevator deflection angle & $-15 \ (deg)$\\
$\delta_e^{max}$ & maximum elevator deflection angle & $30 \ (deg)$\\
\hline
\end{tabular}
\end{table}
\vfill

\end{document}